%% file: main.tex
\documentclass[runningheads]{llncs}

\usepackage[mobile]{eccv}

\usepackage{eccvabbrv}

\usepackage{graphicx}
\usepackage{booktabs}

\usepackage[accsupp]{axessibility}  %

\usepackage[pagebackref,breaklinks,colorlinks,citecolor=eccvblue]{hyperref}

\usepackage{orcidlink}

\usepackage{multirow}
\usepackage{algpseudocode}
\usepackage{algorithm}
\usepackage{xfrac}

\begin{document}

\title{Close, But Not There: Boosting Geographic Distance Sensitivity in Visual Place Recognition}

\titlerunning{Close, But Not There}

\author{Sergio Izquierdo\orcidlink{0000-0002-5639-5035} \and
Javier Civera\orcidlink{0000-0003-1368-1151}}

\authorrunning{S.~Izquierdo and J.~Civera}
\institute{I3A, University of Zaragoza, Spain\\
\email{\{izquierdo, jcivera\}@unizar.es}}

\maketitle

\input{sec/0_abstract}

\input{sec/1_intro}
\input{sec/2_related}

\input{sec/3_analysis}

\input{sec/4_method}

\input{sec/5_experiments}

\input{sec/6_conclusion}

\bibliographystyle{splncs04}
\bibliography{main}
\end{document}

%% file: sec/0_abstract.tex
\begin{abstract}
Visual Place Recognition (VPR) plays a critical role in many localization and mapping pipelines. 
It consists of retrieving the closest sample to a query image, in a certain embedding space, from a database of geotagged references.
The image embedding is learned to effectively describe a place despite variations in visual appearance, viewpoint, and geometric changes.
In this work, we formulate how limitations in the \emph{Geographic Distance Sensitivity} of current VPR embeddings result in a high probability of incorrectly sorting the top-$k$ retrievals, negatively impacting the recall. In order to address this issue in single-stage VPR, we propose a novel mining strategy, \emph{CliqueMining}, that selects positive and negative examples by sampling cliques from a graph of visually similar images. Our approach boosts the sensitivity of VPR embeddings at small distance ranges, significantly improving the state of the art on relevant benchmarks. In particular, we raise recall@1 from 75\% to 82\% in MSLS Challenge, and from 76\% to 90\% in Nordland. Models and code are available at \href{https://github.com/serizba/cliquemining}{https://github.com/serizba/cliquemining}.
  \keywords{Visual Place Recognition \and Image Retrieval}
\end{abstract}

%% file: sec/1_intro.tex
\section{Introduction}
\label{sec:intro}

Visual Place Recognition (VPR) refers to identifying a place from a query image $\mathcal{I}_q \in \mathbb{R}^{w \times h \times 3}$, which boils down to retrieving the $k$ closest images $\{\mathcal{I}_1, \hdots, \mathcal{I}_k\}$ from a database where they are georeferenced. VPR is fundamental in several computer vision applications. It constitutes the first stage of visual localization pipelines by providing a coarse-grain pose that reduces the search space in large image collections. This pose can be later refined by robust geometric fitting from local feature matches~\cite{sarlin2021back,panek2023visual}. It is also essential in visual SLAM, in which it is used to detect loop closures and remove geometric drift~\cite{cadena2016past,campos2021orb}, or as the basis for topological SLAM~\cite{garcia2017hierarchical,doan2019scalable}. 

In VPR pipelines, every RGB image $\mathcal{I}_i$ is typically mapped to a low-\hspace{0pt}dimensional embedding $x_i \in \mathbb{R}^d$ by a deep neural network $f_\theta : \mathcal{I}_i \rightarrow x_i$ that extracts and aggregates visual features that are relevant for the task. The closest samples are retrieved by a nearest-neighbour search using distances in the embedding space $d_i^e = ||x_q - x_i||_2$, which hopefully correspond to the views with smallest geographic distance $d_i^g = ||p_q - p_i||_2$ between them, with $p_i \in \mathbb{R}^3$ standing for the camera position for $\mathcal{I}_i$. The challenge lies on learning the wide variability in the visual appearance of places, caused among others by environmental, weather, seasonal, illumination and viewpoint variability, or dynamic content. Recent years have witnessed significant advances in VPR, driven among others by enhanced network architectures~\cite{ali2023mixvpr, izquierdo2024optimal,ali2022gsv,lu2024towards,zhu2023r2former}, loss functions~\cite{hadsell2006dimensionality,weinberger2005distance,wang2019multi,leyva2023data}, or two-stage re-ranking strategies~\cite{cao2020unifying,hausler2021patch,wang2022transvpr,shen2023structvpr,zhu2023r2former}. %

\begin{figure}[t]
  \centering
    \includegraphics[width=0.99\linewidth]{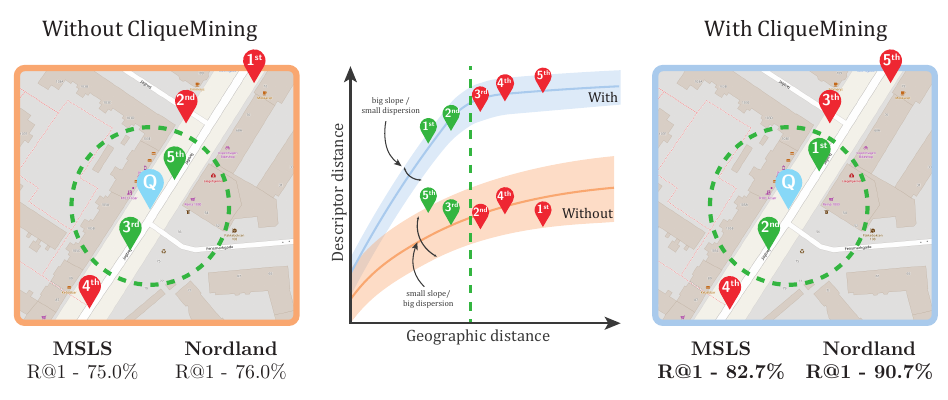}
    \caption{\textbf{Geographic Distance Sensitivity (GDS).} We illustrate a typical case of top-$5$ retrieval without (left) and with (right) our proposed CliqueMining. Note how retrievals on the left are not properly sorted based on geographic distance, impacting the recall for the selected threshold (green circle). We conceptualize this effect as GDS in the central plot, which shows the distribution of descriptor distances against geographic distances. A low slope of the mean (orange line) and a high dispersion (orange area), indicative of low GDS, raise the probability of an incorrect order. To address this, we present CliqueMining, a novel batch selection pipeline that increases the GDS of a model (blue line and area) and produces more correct retrievals.}
    \label{fig:teaser}
\end{figure}\vspace{0.5em}

In this work, we start by analyzing the Geographic Distance Sensitivity (GDS) of VPR embeddings, that can be illustrated by a plot of the distribution of embedding distances $d^e$ vs. geographic distances $d^g$, as in the centre of \cref{fig:teaser}. The plot shows two cases: in orange the distribution a typical VPR pipeline would achieve, and in blue the distribution that would be obtained by a model with enhanced GDS, result of training using our novel CliqueMining, which we will introduce later.
Note how a high variance and a small slope results in a high probability of incorrectly sorting the top-$5$ retrievals. The top-1 retrieval on the left is, as it is written in the title, close but not there. By decreasing the variance and increasing the slope the probability of an incorrect ordering decreases. 

\begin{figure}
  \centering
    \includegraphics[width=0.99\linewidth]{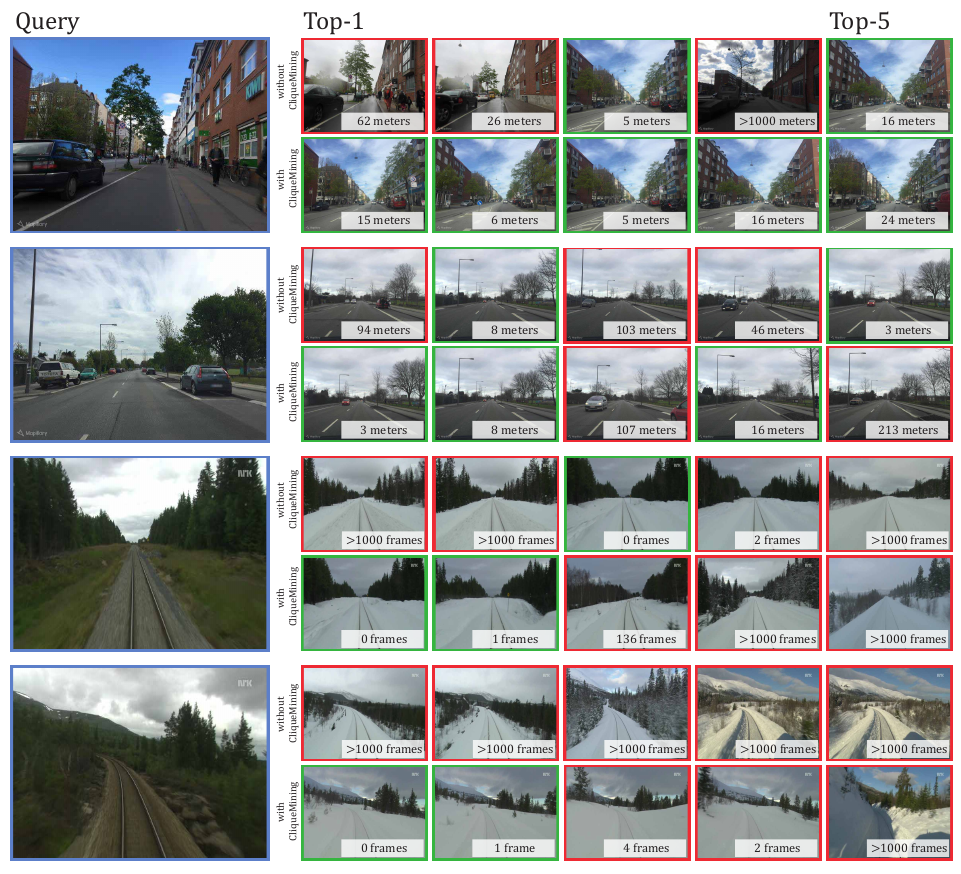}
    \caption{Top-$5$ retrievals for DINOv2-SALAD~\cite{izquierdo2024optimal} without and with our CliqueMining in MSLS~\cite{warburg2020mapillary} and Nordland~\cite{nordland}. Green frames represent correct retrievals and red frames incorrect ones, under the standard $25$-meters (1 frame for Nordland) decision threshold. Our CliqueMining achieves a better sorting of the retrievals with respect to their geographical distance to the query, which positively impacts the recall.}
    \label{fig:topkqualitative}
\end{figure}

\cref{fig:topkqualitative} shows this phenomenon occurring in real datasets when using the state-of-the-art baseline DINOv2 SALAD~\cite{izquierdo2024optimal}. Observe how the top-$5$ retrievals without our CliqueMining in MSLS~\cite{warburg2020mapillary} and Nordland~\cite{nordland} are not properly sorted by real geographic distance.
While two-stage re-ranking approaches might assist in alleviating this, their local feature matching stage come with a prohibitive storage and computational footprint. %
Additionally, recent methods using only global features~\cite{shao2023global,izquierdo2024optimal} already surpass those that involve local features for re-ranking.
Although mining strategies also aim to improve performance by compiling informative batches during training, existing strategies are not specifically tailored to enhance GDS in densely sampled data. %

In addition to analyzing GDS, in this work we propose a novel mining strategy, CliqueMining, explicitly tailored to address it. Our hypothesis is that, in order to boost the GDS, the training batches should include images of highly similar appearance at small distances, that are not explicitly searched for in current mining schemes. We achieve that by organizing our training samples as a graph from which we extract cliques that represent sets of images that are geographically close. Our experiments show that, in in this way, using CliqueMining on top of a baseline model obtains substantial improvements in recall metrics.

%% file: sec/2_related.tex
\section{Related Work}
\label{sec:related}

Early approaches to VPR were mainly based on aggregating handcrafted  features, \eg~\cite{cummins2008fab,jegou2010aggregating,perronnin2010large,galvez2012bags,torii2013visual,arandjelovic2013all}. %
More recent ones have sometimes used deep backbones that were pre-trained in supervised~\cite{sunderhauf2015performance,sunderhauf2015place} and unsupervised~\cite{keetha2023anyloc} setups, showing better generalization and performance. However, the state of the art has been typically represented in the last years by deep models specifically trained or fine-tuned for VPR tasks, \eg, \cite{arandjelovic2016netvlad,radenovic2018fine,berton2022rethinking,berton2023eigenplaces,lu2024towards,izquierdo2024optimal}. For the general perspective and evolution of VPR over the last years, too vast to be fully referenced here, the reader is referred to existing tutorials and surveys focused specifically on VPR~\cite{lowry2015visual,zhang2021visual,masone2021survey,garg2021your,Schubert2023vpr} or on general content-based image retrieval~\cite{chen2022deep}. In the rest of the section, we will only explicitly cite those works that are most related to our contribution.

Overall, training details matter in image retrieval, and are task-specific. Typically, contrastive~\cite{hadsell2006dimensionality} and triplet~\cite{weinberger2005distance} losses are used to train a deep model that maps images into an embedding space, in which similar samples are close together and dissimilar ones are far apart. Although other losses have been proposed in the literature, \eg~\cite{sohn2016advances,wang2017deep, wu2017sampling,yuan2019signal,Deng2018ArcFaceAA,cakir2019deep,Sun2020CircleLA,Boutros2021ElasticFaceEM}, Musgrave \etal~\cite{musgrave2020metric} and Roth \etal~\cite{roth2020revisiting} showed a higher saturation than the indicated in the literature. The particularities of VPR, however, can be leveraged in task-specific losses. For example, Leyva-Vallina \etal~\cite{leyva2023data} grade similarity based on spatial overlap to make losses more informative. Ali-bey \etal~\cite{ali2022gsv} showed that the multi-similarity loss~\cite{wang2019multi} can be effectively used for VPR tasks. They curated a dataset, GSV-Cities, and organized it on sparse places that, combined with the multi-similarity loss led to significant performance gains. As other recent works~\cite{ali2023mixvpr,izquierdo2024optimal}, our contribution builds on top of the multi-similarity loss on GSV-Cities. However, the sparse nature of the GSV-Cities dataset~\cite{ali2022gsv} limits the GDS of the models in densely sampled data, present in many benchmarks~\cite{nordland,warburg2020mapillary}. We argue that densely sampled data is relevant in VPR as it is a prevalent condition in numerous applications, owing to the proliferation of mobile computational platforms capturing video (such as cars, drones, glasses and phones) and the availability of tools to crowdsource and store big data.

Mining informative batches matters as much or even more than the chosen losses~\cite{wu2017sampling}. ``Easy'' samples contribute with small loss values, which may slow down or plateau the training~\cite{musgrave2020metric}. On the other hand, using only ``hard'' samples produces noisy gradients and may overfit or converge to local minima~\cite{radenovic2016cnn,wu2017sampling}, which suggests a sweet spot in mixed strategies~\cite{schroff2015facenet}. As another taxonomy, mining can be done offline after a certain number of iterations~\cite{harwood2017smart,smirnov2017doppelganger,suh2019stochastic}, with high computational costs, or online within each batch~\cite{hermans2017defense,yuan2017hard}. In practice, ``Hard'' negatives samples are typically used, as they are easy to mine and informative~\cite{arandjelovic2016netvlad,xuan2020hard,kalantidis2020hard,warburg2020mapillary}. ``Hard'' positive mining~\cite{simo2014fracking,jin2017learned,ge2020self,lu2023aanet} is more challenging to implement, as it is sometimes caused by occlusions, large scale changes or low overlap, which may be misleading and harm generalization~\cite{radenovic2018fine}. Wang \etal~\cite{wang2019multi} generalizes sampling schemes by weighting pairs in the multi-similarity loss according to their embedding distance. None of the mining approaches in the literature, however, addresses GDS aspects as we do with our CliqueMining.

%% file: sec/3_analysis.tex
\section{Geographic Distance Sensitivity in VPR}
\label{sec:sensitivity}

As already said, \cref{fig:topkqualitative} shows examples of DINOv2 SALAD~\cite{izquierdo2024optimal} retrievals on MSLS Train~\cite{warburg2020mapillary} and Nordland~\cite{nordland}. Although the recall@1 for these specific queries is zero, dismissing the model's performance as entirely inaccurate would be unfair. 
Within the top-5 retrievals, some predictions are indeed correct, and most incorrect predictions are relatively close to the decision threshold. These examples uncover a common issue in VPR models:  their inability to finely discriminate between similar viewpoints. Note how our novel CliqueMining, that we will describe in next sections, discriminates better for this particular case.

We explain this phenomenon using the concept of Geographic Distance Sensitivity (GDS), \emph{i.e.}, the model's ability to assign smaller descriptor distances to pairs of images that are geographically closer. VPR models should have a high GDS, that is, they should produce descriptors that maximize the probability $P(d_i^e < d_j^e \ | \  d_i^g < d_j^g)$. Seeking for a high GDS requires two desiderata to hold. 

(i) The expected value of the descriptor distance of a pair should be smaller than that of a pair geographically further from the query $\mathbb{E}[d_i^e - d_j^e \ | \ d_i^g < d_j^g] < 0$. 

(ii) The dispersion of descriptor distances conditioned on a certain geographic distance should be as small as possible $\mathbb{E}[(d_i^e - \mathbb{E}[d_i^e \ | \ d_i^g])^2 \ | \ d_i^g ] \rightarrow 0$.

Failing to achieve these two leads to a high probability of retrieving an incorrect order of candidates. We hypothesize that VPR models struggle to precisely rank between closely spaced locations due to their limited GDS at small distance ranges. This is because current training pipelines are effective at achieving highly invariant representations that encode viewpoints coarsely, but not at learning the subtle cues to disambiguate between close frames.

\begin{figure}[t!]
  \centering
  \begin{subfigure}{0.49\linewidth}
    \includegraphics[width=\linewidth]{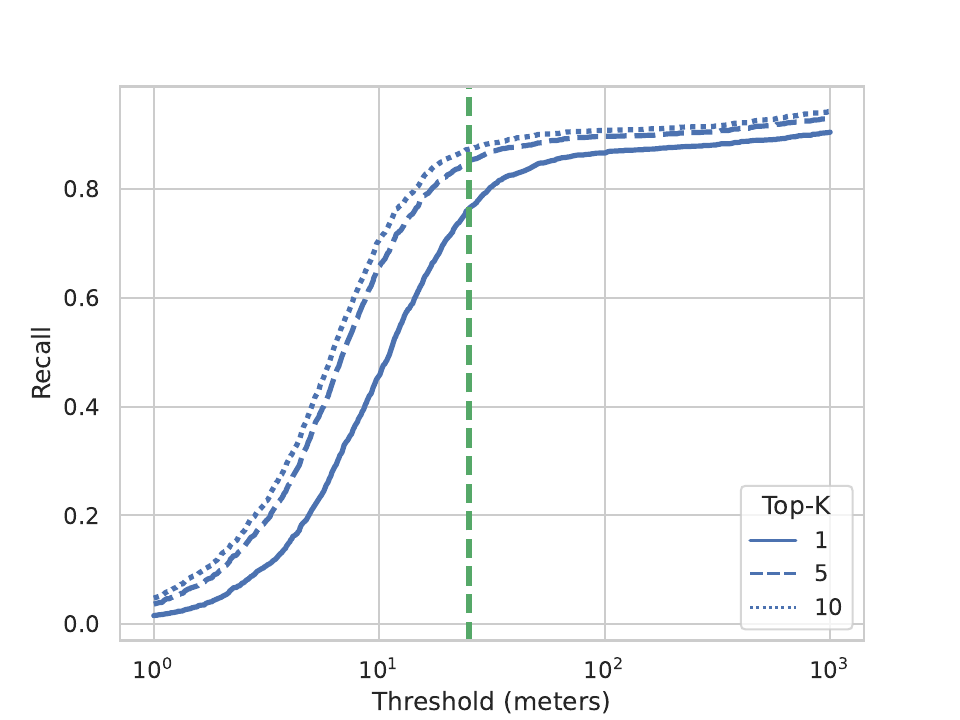}
    \caption{MSLS Train (Val Subset)}
    \label{fig:recall_curves_msls_train}
  \end{subfigure}
  \hfill
  \begin{subfigure}{0.49\linewidth}
    \includegraphics[width=\linewidth]{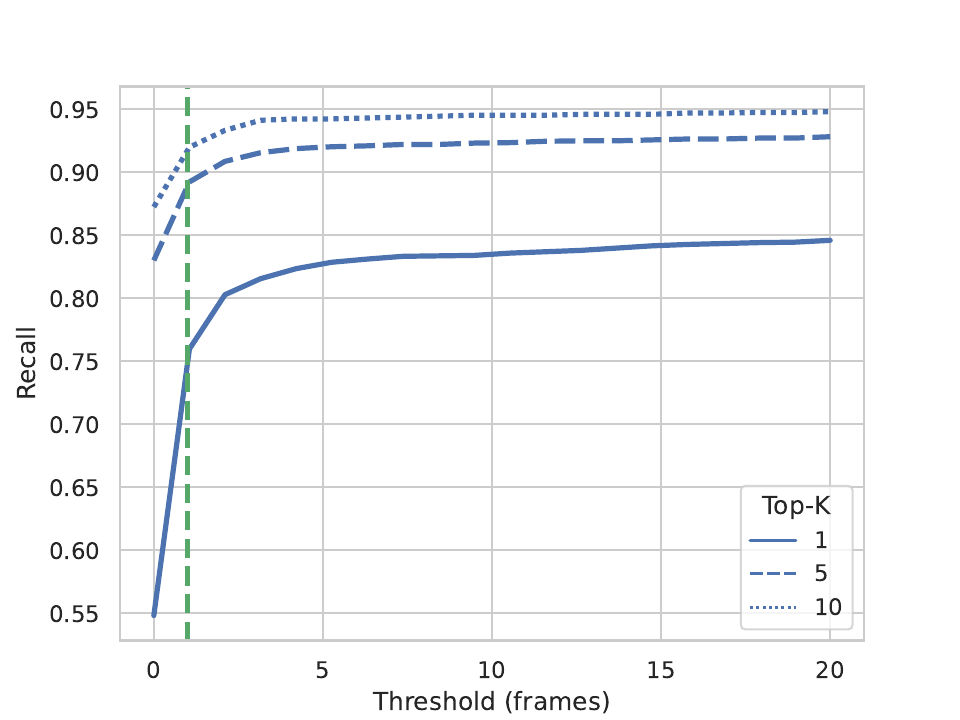}
    \caption{Nordland}
    \label{fig:recall_curves_nordland}
  \end{subfigure}
  \caption{\textbf{Recall@K vs. decision threshold} on MSLS Train (val) and Nordland for DINOv2-SALAD~\cite{izquierdo2024optimal} without CliqueMining. Observe how the steep curve around the decision threshold (green dashed line) indicates a significant number of closely retrieved images. Boosting the GDS of a model would alleviate this, increasing its recall. }
  \label{fig:recall_curves}
\end{figure}

This effect can be further assessed in \cref{fig:recall_curves}, which shows the top-$\{1,5,10\}$ recall of the baseline DINOv2 SALAD for different threshold values. The vertical green dashed lines represent the typical thresholds of $25$ meters and $1$ frame used in MSLS and Nordland. Note how the recall, specially the recall@1, keeps increasing for slightly larger values than the $25$ meters and $1$ frame thresholds. This indicates that a significant fraction of false negatives is very close to the decision threshold, which lowers the recall. 
With our novel CliqueMining strategy, detailed in next section, the reader will assess how we are able increase the GDS for small ranges (Fig. \ref{fig:distances_plot}) and consequently improve recall metrics, as we will show in the experimental results

%% file: sec/4_method.tex
\section{CliqueMining}

\begin{figure}[t]
  \centering
    \includegraphics[width=0.99\linewidth]{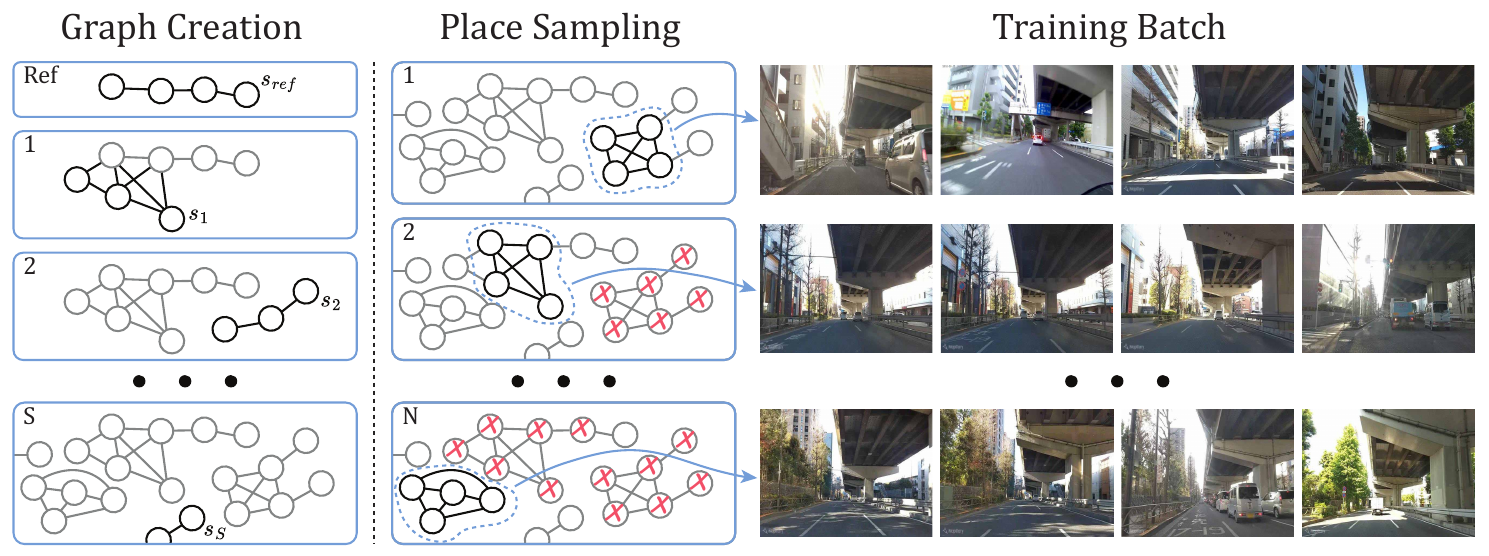}
    \caption{\textbf{Overview of CliqueMining.} First, we create a graph of candidates by sampling a set of sequences $\{s_1, \dots, s_S\}$ that are similar to a reference one $s_{ref}$ (left). We then sample places by finding cliques within the graph (center). Observe that the resulting batches contain very similar looking places, which boost the GDS (right).}
    \label{fig:method}
\end{figure}

Our novel mining strategy, CliqueMining, selects challenging batches according to geographic and descriptor similarity criteria, alleviating the GDS issues identified in Section~\ref{sec:sensitivity}. \cref{fig:method} shows an overview of our method. To effectively mine a challenging batch, we first build a graph of image candidates (\cref{sec:graph_creation}) and sample places from it (\cref{sec:graph_sampling}). Finally, we select challenging pairs and train the network using the Multi-Similarity loss (\cref{sec:training}). 

\subsection{Graph Creation}
\label{sec:graph_creation}

In contrast with the sparse nature of viewpoint sampling in GSV-Cities~\cite{ali2022gsv}, we propose to use denser batches, with higher spatial continuity, so the the network also learns the subtle changes resulting from small camera motion. To effectively mine such challenging batches, we first create a graph, $G=(V, E)$, representing a cluster of candidates. Vertices from this graph, $v_i\in V$, are frames from sequences with very similar appearance, and two vertices, $v_i$ and $v_j$, are connected by an edge $e_{ij}\in E$ if both frames lie within a given distance threshold in meters, $\tau$. 
\begin{equation}
E = \{e_{ij} \ | \ d(v_i, v_j) < \tau , \ \forall v_i, v_j \in V\}
\end{equation}
To populate the graph, we consider all image sequences as defined in the MSLS training set, as our place-based batches do not require a split between query and database images. We start by sampling a reference sequence from a city, $s_{ref}$, and subsequently, sampling $S$ more different sequences, $\{s_1, \dots, s_S\}$ based on their similarity with $s_{ref}$. For computational efficiency, we determine the similarity between two sequences by only comparing the descriptors of their respective central frames. We incorporate every frame from these sequences into the graph, which ensures the presence of adjacent frames within the batches. Edges are determined by the Universal Transverse Mercator (UTM) locations of each frame. \Cref{alg:graph_creation} summarizes this process. 

\algdef{SE}[REPEATN]{RepeatN}{EndRepeatN}[1]{\algorithmicrepeat\ #1 \textbf{times}}{\algorithmicend}
\begin{algorithm}
    \caption{Graph creation.}
    \label{alg:graph_creation}
    \begin{algorithmic}
        \State Initialize $G=(V, E)$ as empty graph
        \State Sample $city$
        \State $V \gets \{v_i | v_i \in s_{ref}\},~ s_{ref} \sim \{s | s \in city\}$ 
        \RepeatN{S}
            \State $s \sim P(s|s_{ref}) \propto sim(s, s_{ref})$
            \State $V \gets V \cup \{v_j | v_j \in s\}$
        \EndRepeatN
        \State $E \gets \{e_{ij} | d(v_i, v_j) < \tau, \forall v_i, v_j \in V\}$
\end{algorithmic}
\end{algorithm}

\subsection{Place Sampling}
\label{sec:graph_sampling}

To construct a single batch, we start from the graph of candidates $G$, generated as explained in \cref{sec:graph_creation}. $G$ is a convenient representation for place sampling, as it facilitates the identification of distinct viewpoints yet of highly similar appearance, and labels are easily assigned based on connectivity. In our pipeline, we mine batches of $N$ places, each place defined as a set of $K$ images, where each image is within a range $\tau$ of each other. Sampling a place is equivalent to finding a clique, $C$, within $G$
\begin{equation}
C \sim \{C \ | \  \forall v_i, v_j \in C, \  e_{ij} \in E, \  C \subseteq V, \ |C| = K\}.
\end{equation}
Thus, to compile a batch of $N$ places, we iteratively extract $N$ cliques from $G$. After finding each clique, all its frames, as well as their connected vertices are removed from $G$. This prevents overlap in subsequent cliques, ensuring that each sampled place is at least $\tau$ meters from each other. In the uncommon case of exhausting all cliques in $G$, we create a new graph starting from a new $s_{ref}$ and continue the process. The resulting batches, an example of them shown in~\cref{fig:method}, showcase highly similar yet far apart images, illustrating the effectiveness of our sampling to create difficult batches. \Cref{alg:graph_sampling} gives an overview of the sampling procedure.

\begin{algorithm}
    \caption{Graph sampling.}
    \label{alg:graph_sampling}
    \begin{algorithmic}
        \State Input: Graph $G=(V, E)$
        \State Initialize empty batch of images $B$
        \State Initialize empty batch of labels $L$
        \ForAll{$n$ from $1$ to $N$}
            \State Sample clique $C \sim \{C \ | \  \forall v_i, v_j \in C, \  e_{ij} \in E, \  C \subseteq V, \ |C| = K\}$
            \State $B \gets B \cup C$
            \State $L \gets L \cup \{n\}$
            \State $G \gets G - \{v_i \cup Adj(v_i) | v_i \in C\}$
        \EndFor
\end{algorithmic}
\end{algorithm}

\subsection{Training Pipeline}
\label{sec:training}

In practice, we mine a large set of batches offline and once, as described in \cref{sec:graph_creation} and \cref{sec:graph_sampling}, and use them during all epochs. To do this, we use the embeddings from a model pre-trained without CliqueMining.
Most mining strategies are typically updated every few iterations. However, this increases the computational overhead, and for our CliqueMining we did not observe any improvement by updating the batches.

In order to smooth the gradients from our hard training images, we combine them with images from GSV-Cities. In this manner, we include per batch half of the images from our CliqueMining and half from GSV-Cities, so the network can learn both the fine-grain GDS and the sparse discriminative capabilities from GSV-Cities.

As we use the Multi-Similarity (MS) loss~\cite{wang2019multi}, during training we use their online selection method for weighted negative and positive pairs. A negative pair, $\{x_i, x_j\}$, is selected from a batch if its distance is lower than the hardest positive pair plus a margin, $\epsilon$,
\begin{equation}
	||x_i - x_j||_2 \ < \ \max_{d_{ik}^e<\tau}{||x_i - x_k||_2+\epsilon},
    \label{eq:mine_neg}
\end{equation}
and, conversely, a positive pair is selected when
\begin{equation}
	||x_i - x_j||_2 \ > \ \min_{d_{ik}^e\geq\tau}{||x_i - x_k||_2-\epsilon}.
    \label{eq:mine_pos}
\end{equation}

%% file: sec/5_experiments.tex
\section{Experiments}
\label{sec:experiments}

In this section, we re-train state-of-the-art VPR baseline models using our proposed CliqueMining. Evaluation on various benchmarks showcases the increased discriminative capacity of the models. In the following, we describe the implementation details, benchmarks used, quantitative and qualitative results, as well as ablation studies.

\subsection{Implementation Details}

We use CliqueMining with the recent DINOv2 SALAD~\cite{izquierdo2024optimal}, the current state-of-the-art VPR model as well as on MixVPR~\cite{ali2023mixvpr}, a recent model with competitive performance. For each of them, we use their codebase and rigorously follow their training pipelines and hyperparameters. We use batches of size $60$ in DINOv2 SALAD and $120$ in MixVPR, where half of the places come from our pipeline and the other half from GSV-Cities. We create a new graph for every batch. We start by sampling $s_{ref}$ from the set of existing sequences.
We then sample $S=15$ sequences from the same city based on the descriptor similarity of their central frames.
Edges are assigned with $\tau=25$. Cliques are searched using the NetworkX library\footnote{\url{https://networkx.org/}} using the unrolled algorithm by Tomita \etal~\cite{tomita2006worst}. We create offline a large collection of $4000$ batch examples before starting the training, and at every iteration, we randomly select one of those. To create the batches we use all the non panoramic images in the MSLS Training set. For the ablation studies we divided this dataset in val and train subsets, setting Melbourne, Toronto, Paris, Amman, Nairobi and Austin for val and the rest 16 cities for train.

\subsection{Results}

\begin{table}
    \centering
    \caption{\textbf{Comparison against single-stage baselines and SelaVPR as representative of two-stage baselines.} Observe the significant increase in the recall in MSLS and Nordland when using CliqueMining (CM). Both are the less saturated datasets, hence with most room for improvement, and the most densely sampled, which is the case our novel CliqueMining is tailored for.}
    \resizebox{\linewidth}{!}{
    \begin{tabular}{l c c c c c c c c c c c c c c c c}
    \hline

    \hline
    \multirow{2}{*}{Method} & \multicolumn{3}{c}{NordLand}&&\multicolumn{3}{c}{MSLS Challenge}&&\multicolumn{3}{c}{MSLS Val}&&\multicolumn{3}{c}{Pitts250k-test}\\ %
    \cline{2-4} \cline{6-8} \cline{10-12} \cline{14-16} %
    & R@1 & R@5 & R@10 && R@1 & R@5 & R@10 && R@1 & R@5 & R@10 && R@1 & R@5 & R@10 \\ %
    \hline
        NetVLAD~\cite{arandjelovic2016netvlad}    & 32.6 & 47.1 & 53.3 && 35.1 & 47.4 & 51.7 && 82.6 & 89.6 & 92.0 && 90.5 & 96.2 & 97.4\\ %
        GeM~\cite{radenovic2018fine}              & 21.6 & 37.3 & 44.2 && 49.7 & 64.2 & 67.0 && 78.2 & 86.6 & 89.6 && 87.0 & 94.4 & 96.3\\ %
        CosPlace~\cite{berton2022rethinking}      & 52.9 & 69.0 & 75.0 && 67.5 & 78.1 & 81.3 && 87.6 & 93.8 & 94.9 && 92.3 & 97.4 & 98.4\\% && 79.6 & 90.4 & 92.8\\
        MixVPR~\cite{ali2023mixvpr}               & 58.4 & 74.6 & 80.0 && 64.0 & 75.9 & 80.6 && 88.0 & 92.7 & 94.6 && 94.6 & 98.3 & 99.0\\% && 85.2 & 92.1 & 94.6\\
        EigenPlaces~\cite{berton2023eigenplaces}  & 54.4 & 68.8 & 74.1 && 67.4 & 77.1 & 81.7 && 89.3 & 93.7 & 95.0 && 94.1 & 98.0 & 98.7\\% && 69.9 & 82.9 & 87.6\\
        SelaVPR (global)~\cite{lu2024towards}     & 47.2 & 66.6 & 74.1 && 69.6 & 86.9 & 90.1 && 87.7 & 95.8 & 96.6 && 92.7 & 98.0 & 98.9 \\%&& & &\\
        SelaVPR (re-ranking)~\cite{lu2024towards} & 60.0 & 75.7 & 79.6 && 73.5 & 87.5 & 90.6 && 90.8 & 96.4 & 97.2 && \textbf{95.7} & \textbf{98.8} & 99.2 \\%&& & &\\
        DINOv2 SALAD~\cite{izquierdo2024optimal}  & 76.0 & 89.2 & 92.0 && 75.0 & 88.8 & 91.3 && 92.2 & 96.4 & 97.0 && 95.1 & 98.5 & 99.1 \\%&& \textbf{92.1} & \textbf{96.2} & \textbf{96.5}\\
    \hline
        {MixVPR~\cite{ali2023mixvpr} CM}   & 69.6 & 80.7 & 83.5 && 65.6 & 77.1 & 79.2 && 88.8 & 93.9 & 94.6 && 91.8 & 96.7 & 98.1 \\%&& 67.0 & 78.9 & 82.2\\
        {DINOv2 SALAD~\cite{izquierdo2024optimal} CM} &\textbf{90.7} & \textbf{96.6} & \textbf{97.5} && \textbf{82.7} & \textbf{91.2} & \textbf{92.7} && \textbf{94.2} & \textbf{97.2} & \textbf{97.4} && 95.2 & \textbf{98.8} & \textbf{99.3} \\%&& 89.3 & 94.7 & 95.9\\
    \hline

    \hline
    \end{tabular}
    }
\label{tab:main}
\end{table}

We evaluate the effect of our CliqueMining by comparing the performance of two recent high-performing models, DINOv2 SALAD~\cite{izquierdo2024optimal} and MixVPR~\cite{ali2023mixvpr}, with and without it at training time. We also benchmarked these against classic methods, namely NetVLAD~\cite{arandjelovic2016netvlad} and GeM~\cite{radenovic2018fine}, and recent performant baselines, specifically CosPlace~\cite{berton2022rethinking}, EigenPlace~\cite{berton2023eigenplaces}, and SelaVPR~\cite{lu2024towards}. Additionally, we include in the comparison results of SelaVPR~\cite{lu2024towards} with re-ranking, as it is the current state of the art among two-stage techniques.

We report results on standard evaluation datasets. \textbf{Nordland}~\cite{nordland} is a continuous video sequence taken from a train traveling through Norway across different seasons. The difficulty of this dataset arises from the substantial appearance differences between query (summer) and reference (winter), as well as the dense temporal sampling. \textbf{MSLS Challenge and Validation}~\cite{warburg2020mapillary} is a large and dense collection of dashcam images recorded in cities around the globe. The various seasonals, time, and environmental changes depicted make it one of the least saturated datasets in VPR. \textbf{Pittsburgh-250k}~\cite{torii2013visual} is known for its significant viewpoint changes, but current pipelines have highly saturated performance.

As previous works, we report recall@$\{1,5,10\}$, which measures the rate of correct predictions among the top-$\{1,5,10\}$ retrieved images. An image is considered correct if it lies within a 25 meters-radius circle from the query, or at most one frame apart for the Nordland dataset. Results are reported on \Cref{tab:main}.

On Nordland, training with our CliqueMining significantly improves both DINOv2 SALAD and MixVPR, obtaining, for the first time, a recall@1 bigger than 90\% (+14.7\% over the closest baseline). This milestone highlights how our hard batches help in boosting the network's GDS. This is a crucial aspect in Nordland, where the high similarity between video frames and the strict one-frame distance threshold need outstanding sensitivity. Note that CliqueMining also improves significantly the recall rates for MixVPR.

On MSLS Challenge and Validation, our CliqueMining with the DINOv2 SALAD architecture improves over all previously reported results. The improvement is most notable on the Challenge, where CliqueMining raises +7.7\% the recall@1. 
While training on the MSLS Train dataset contributes to these results, it is noteworthy that SelaVPR, which also trains on MSLS, does not achieve a comparable performance, even with re-ranking.
The effect of CliqueMining on MixVPR is dimmer, although it also improves over the baseline without it. We argue that its global aggregation smooths out local details, which are critical for raising the GDS.

On Pittsburgh-250k, our pipeline obtains a slight improvement over the baseline DINOv2 SALAD and obtains comparable performance to SelaVPR with re-ranking. We outperform SelaVPR without re-ranking, which is a more comparable baseline. Note, in any case, that SelaVPR is fine-tuned on Pittsburgh30k before testing on Pittsburgh250k, while ours was trained in GSV-Cities and MSLS. MixVPR with CliqueMining downgrades performance. Training on MSLS data, where almost all images are forward-facing, has a small impact on Pittsburgh250k, which exhibits substantial viewpoint variability.

Note how we sorted the datasets in \cref{tab:main} from more to less image density, and how this also sorted naturally the recall@1 gains of CliqueMining from bigger to smaller. This supports our observation that GDS issues are more relevant the higher the image density, and that CliqueMining is able to improve them. From these results we can also conclude that a substantial part of the challenge in the less saturated VPR datasets (Nordland and MSLS) is associated to GDS issues, which is a relevant insight.

\begin{figure}
  \centering
  \begin{subfigure}{0.49\linewidth}
    \includegraphics[width=\linewidth]{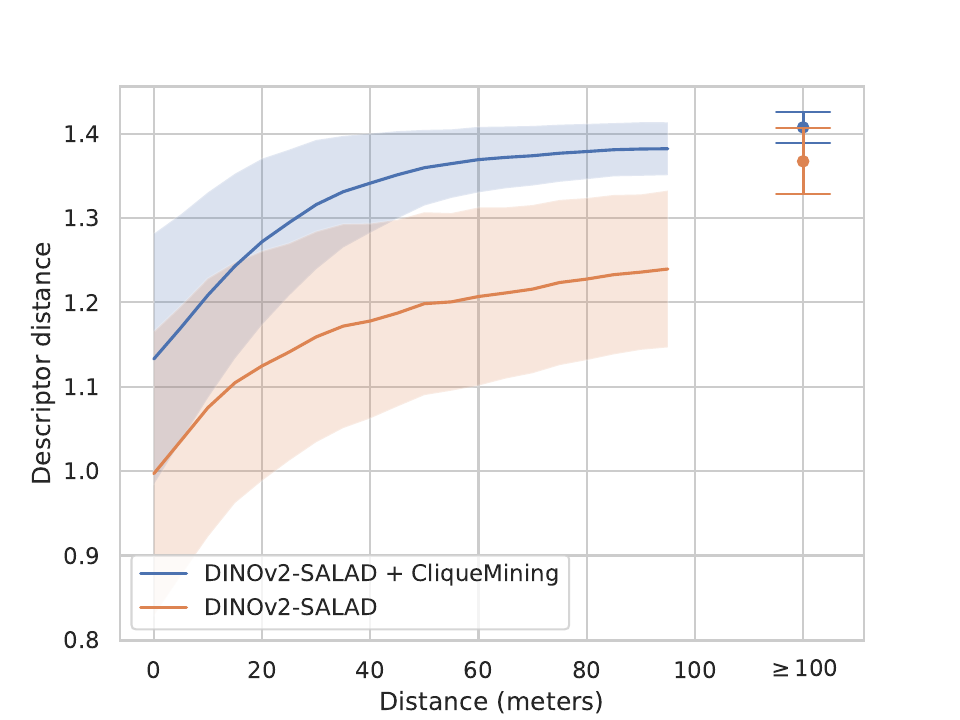}
    \caption{MSLS Train (Val Subset)}
    \label{fig:distances_plot_msls_train}
  \end{subfigure}
  \hfill
  \begin{subfigure}{0.49\linewidth}
    \includegraphics[width=\linewidth]{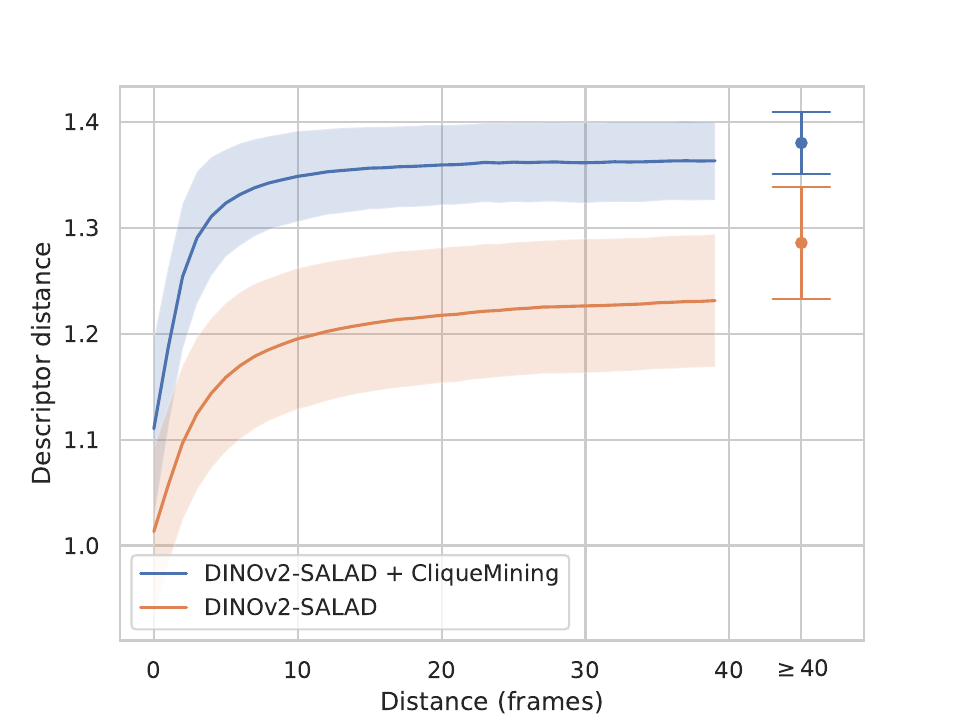}
    \caption{Nordland}
    \label{fig:distances_plot_nordland}
  \end{subfigure}
  \caption{\textbf{Mean $\pm$ standard deviation of descriptor distances against geographic distances, without and with CliqueMining. }Our Clique Mining boosts the geographic local sensitivity for small geographic distances, and flattens it for large distances. This results in higher discriminativity around the decision threshold and better metrics. Note the cut in distances and values for high distances aggregated at the right part.}
  \label{fig:distances_plot}
\end{figure}

Observe in \cref{fig:distances_plot} the effect of CliqueMining on the GDS of the DINOv2-SALAD model~\cite{izquierdo2024optimal} in MSLS and Nordland, as a plot of the distribution of the pairwise descriptor distances for different geographic distances. As sought, the GDS is highly boosted (steep curve and low dispersion) by CliqueMining for close geographic distances. Observe the similarity of this result with the illustrative graph in \cref{fig:teaser}. Although not specifically tailored for, CliqueMining also reduces the dispersion for large distances, probably due to leveraging batches with more informative gradients. This enables the model to correctly sort candidates that are near, and still discriminate from those too far apart.

We finally remark the low computational footprint of our CliqueMining. CliqueMining is a mining strategy for training, and hence does not increase at all the computational footprint at inference. This is in contrast to two-stage methods, that increase it by a factor of several orders of magnitude. Additionally, the overhead is modest at training. Our ablations shows that the graph creation only needs to be done once before training, and there is no benefit in updating it. In total, the computational overhead of CliqueMining roughly amounts to only 20\% of the total training time in our experiments.

\subsection{Ablation Study}

\begin{table}[ht]
    \centering
     \caption{\textbf{Ablations.} First row shows the recall for the base DINOv2-SALAD model. Note in the next three rows that random sampling based on sequence similarity outperforms slightly a determininistic sampling of the most similar ones and some more a uniform random sampling. The MS mining also plays a role in the performance. Note how training on GSV-Cities + MSLS w/o CliqueMining, which accounts for the domain change effect, still underperforms at R@1. Finally, note that recomputing cliques every epoch gives metrics that are similar to computing them only once.
     }
    {
    \begin{tabular}{l cc cc c}
    \hline

    \hline
    \multirow{2}{*}{Method} & \multicolumn{5}{c}{MSLS Train (Val Subset)}\\ 
    \cline{2-6}
    & R@1 && R@5 && R@10 \\
    \hline DINOv2 SALAD~\cite{izquierdo2024optimal} & 76.3 && 85.1 && 87.3\\
    \hline
        Most-similar & 81.61 $\pm$ 0.50 && 89.43 $\pm$ 0.53 && 91.02 $\pm$ 0.53 \\
        Weighted random sampling \ & 81.98 $\pm$ 0.75 && 89.72 $\pm$ 0.22 && 91.12 $\pm$ 0.14 \\
        Uniform random sampling  & 80.40 $\pm$ 0.70 && 87.33 $\pm$ 0.88 && 88.95 $\pm$ 0.70 \\
    \hline
         W/o MS mining & 76.87 $\pm$ 0.46 && 83.92 $\pm$ 0.60 && 86.05 $\pm$ 0.76 \\
    \hline
        Naïve GSV-Cities + MSLS & 79.96 $\pm$ 0.46 && 89.71 $\pm$ 0.32 && 91.80 $\pm$ 0.30 \\
    \hline Recompute Cliques & 81.96 $\pm$ 0.59 && 89.64 $\pm$ 0.54 && 91.28 $\pm$ 0.39 \\
    \hline
    
    \hline
    & \multicolumn{5}{c}{Nordland}\\ 
    \cline{2-6}
    & R@1 && R@5 && R@10 \\
    \hline
    DINOv2 SALAD~\cite{izquierdo2024optimal} & 76.0 && 89.2 && 92.0\\ 
    \hline
    Weighted random sampling & 88.22 $\pm$ 0.99 && 95.22 $\pm$ 0.45 && 96.52 $\pm$ 0.38 \\
    \hline
    Naïve GSV-Cities + MSLS & 68.27 $\pm$ 5.47 && 82.92 $\pm$ 4.99 && 86.81 $\pm$ 4.36 \\
    \hline
    
    \hline
    \end{tabular}}
\label{tab:sampling}
\end{table}

We conduct evaluations with different configurations of CliqueMining to assess the importance of its different components. We base all our ablation studies on the DINOv2 SALAD baseline.

\textbf{CliqueMining or training on more data.} One of the key contributions of this work is to train state-of-the-art models on a combination of GSV-Cities and MSLS. This raises the question of whether the observed improvements result from training with more data or from CliqueMining. To evaluate this, we re-train DINOv2 SALAD on a combination of GSV-Cities + MSLS without CliqueMining. Thus, batches from MSLS are organized in triplets as usually done in the literature. \cref{tab:sampling} shows how, although training on MSLS slightly increases performance, using CliqueMining produces the best results, specially for R@1. We also report, for this ablation, results on Nordland which show more pronounced differences with CliqueMining. This suggest that naïvely training on more data brings limited improvements. CliqueMining creates challenging batches that improve the sensitivity of the model and its recall. Besides, CliqueMining organizes the images in places, so every image can simultaneously act as an anchor, positive or negative, increasing the number of pairwise relations on a batch. 

\begin{figure}[h!]
  \centering
    \includegraphics[width=0.6\linewidth]{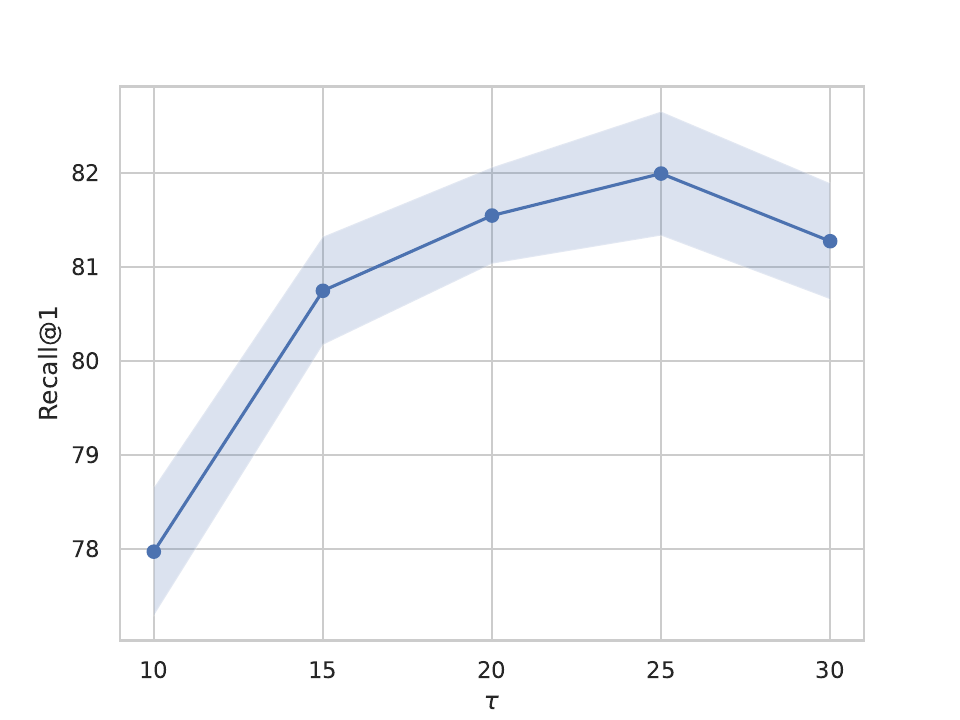}
    \caption{\textbf{Recall@1 on MSLS Train (val) for different values of $\boldsymbol{\tau}$}. Note how, reasonably, $\tau=25$ meters, which is equal to the decision threshold, is the best value.}
    \label{fig:tau}
\end{figure}

\textbf{Geographic distance threshold $\tau$.}
We tested the effect of the $\tau$ values in the range $10$-$30$. As shown in \cref{fig:tau}, using the typical decision threshold value $\tau=25$ achieves the best performance.

\textbf{Multi-Similarity (MS) mining}. We built our CliqueMining on top of \cite{ali2022gsv}, keeping its online mining (\cref{eq:mine_neg,eq:mine_pos}). Deactivating it, keeping only our CliqueMining, has a detrimental effect (see \cref{tab:sampling}), which indicates that both mining strategies are compatible.

\textbf{Sequence sampling.} We evaluate the effect of different sampling strategies to obtain $\{s_1, \dots, s_S\}$ during the graph creation. We specifically try a weighted sampling according to similarity, selecting the top $S$ most similar sequences, or randomly. \cref{tab:sampling} shows that all three sampling strategies obtain very similar results, but using the most similar sequences produces the best. We argue that the online mining from \cref{eq:mine_neg,eq:mine_pos} reduces the actual differences between the used selection criteria, as it will further select the hardest pairs. Besides, given the length of some of the sequences, more than one clique might be sampled from the same sequence, reducing the need to find other similar ones.

\textbf{Updating the mining every epoch.} Commonly done in literature, updating the mining after every epoch using the recently updated weights can provide some benefits to performance. As shown in \cref{tab:sampling}, obtained recalls are comparable, and computing the mining after every epoch is computationally expensive.

%% file: sec/6_conclusion.tex
\section{Limitations}
\label{sec:limitations}

\begin{figure}
  \centering
  \begin{subfigure}{0.49\linewidth}
    \includegraphics[width=\linewidth]{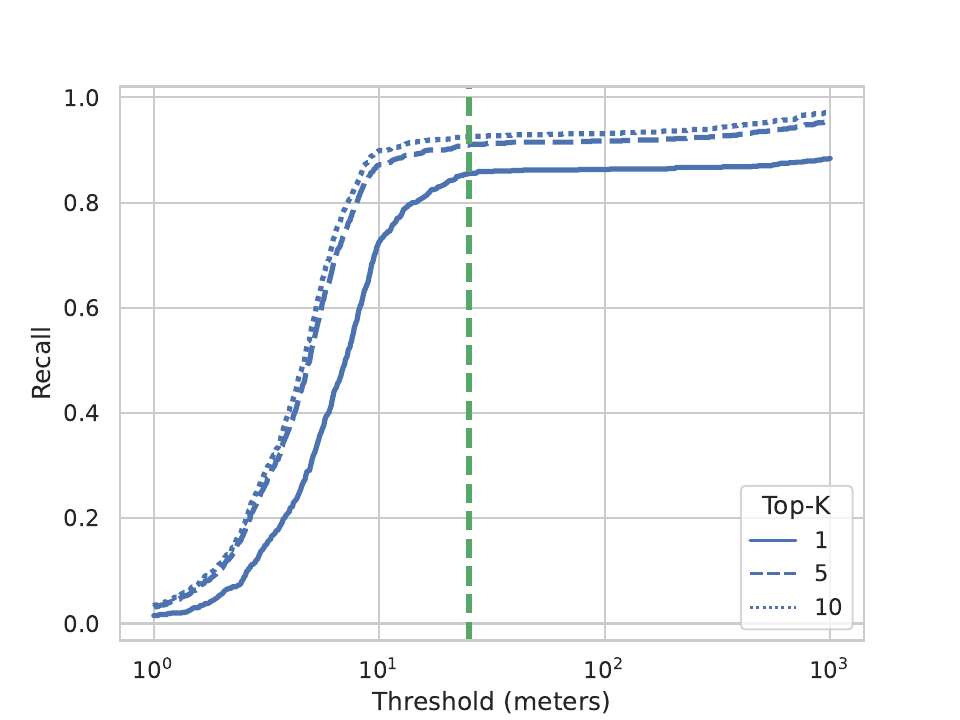}
    \caption{SF-XL test v1}
    \label{fig:recall_curves_sfxlv1}
  \end{subfigure}
  \hfill
  \begin{subfigure}{0.49\linewidth}
    \includegraphics[width=\linewidth]{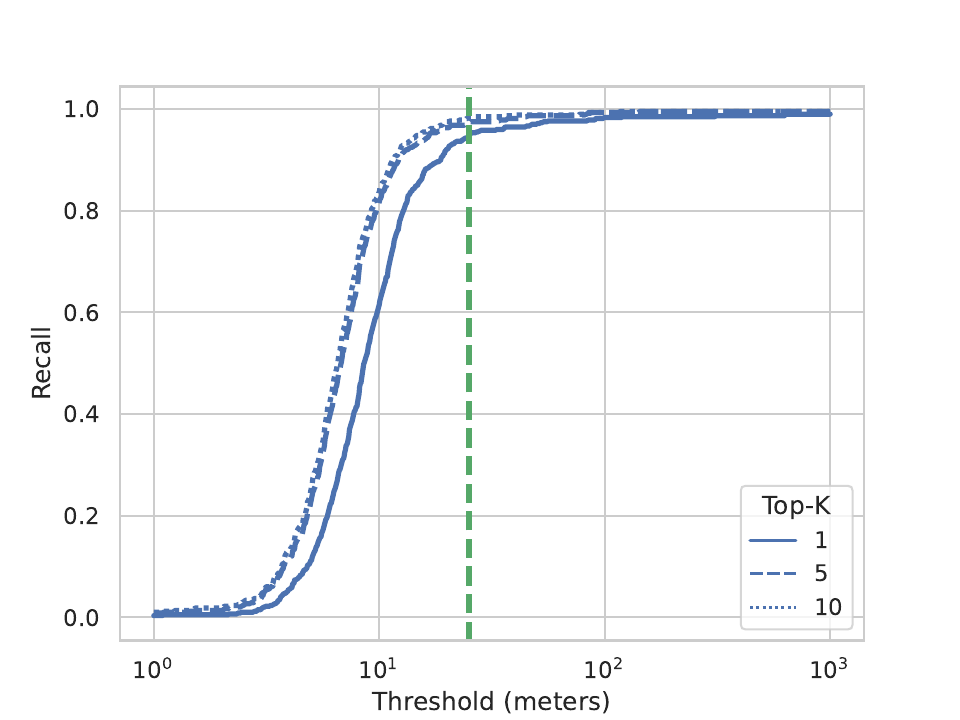}
    \caption{SF-XL test v2}
    \label{fig:recall_curves_sfxlv2}
  \end{subfigure}
  \caption{\textbf{Recall@K vs. decision threshold on SF-XL} for DINOv2-SALAD~\cite{izquierdo2024optimal} without CliqueMining. Observe how the recall curves are almost flat beyond the decision threshold (green dashed line), indicating a low false negative rate due to limited GDS. Compare it against the recall curves in MSLS and Nordland in \cref{fig:recall_curves}. In this case, enhancing the GDS will not result in better metrics.}
  \label{fig:recall_curves_sfxl}
\end{figure}

The main limitation of CliqueMining is that it is specifically tailored for VPR, and hence it will not be of use for general image retrieval. In addition, CliqueMining addresses GDS issues, that are mostly relevant for places that are densely sampled with images. We already reported in \cref{tab:main} the diminishing returns as the sampling density decreases in the benchmarks we used. However, as we motivated in \cref{sec:related}, this limitation is softened by the wide range of potential use cases falling into this condition, and also by the remarkable boost in recall@1 in the most dense sampling cases ($+14.7\%$ for Nordland).

Additionally to the above, our CliqueMining is strongly dependent on the existence of GDS issues. 
Even if the dataset is densely sampled, there could be a lack of GDS issues, as when viewpoint changes account for the majority of variations. In this cases, the model fails to retrieve close samples, and therefore CliqueMining would not positively impact its recall.
We observed this in the recent {SF-XL}~\cite{berton2022rethinking}, a massive dataset of images from San Francisco, often used to test VPR at scale. \cref{fig:recall_curves_sfxl} characterizes the recall in this dataset against the decision threshold. Observe how, in contrast to \cref{fig:recall_curves}, the recall is almost flat in the region immediately after the decision threshold. Enhancing the GDS is not expected to have any effect in this dataset, as the rate of false negatives due to this reason is very small. Even if this is a limitation, we would argue in our favour that every mining strategy is strongly dependent on the data, but in the case of our CliqueMining we have characterized the conditions in which it should or should not offer an improvement.

\section{Conclusions}
\label{sec:conclusions}

In this paper we have identified, formulated and analyzed deficiencies in the Geographic Distance Sensitivity (GDS) of current VPR models. Specifically, we found that they struggle to correlate descriptors and geographic distances for close range views. Based on that, we propose CliqueMining, a tailored batch sampling that selects challenging visually similar places at close ranges, and in particular around the decision threshold. CliqueMining forces the model to incorporate a finer grading of the geographic distances in the embedding. Mining such hard batches is equivalent to finding cliques in a graph of similar image sequences where connectivity represents spatial proximity. Our evaluation of two recent models with and without CliqueMining confirms a boost in the GDS which in turn also boosts the recall. The boost is substantial on densely sampled and unsaturated benchmarks like MSLS Challenge or Nordland, where training with CliqueMining brings unprecedented results.

%% file: main.bbl
\begin{thebibliography}{10}
\providecommand{\url}[1]{\texttt{#1}}
\providecommand{\urlprefix}{URL }
\providecommand{\doi}[1]{https://doi.org/#1}

\bibitem{ali2022gsv}
Ali-bey, A., Chaib-draa, B., Gigu{\`e}re, P.: Gsv-cities: Toward appropriate
  supervised visual place recognition. Neurocomputing  \textbf{513},  194--203
  (2022)

\bibitem{ali2023mixvpr}
Ali-Bey, A., Chaib-Draa, B., Giguere, P.: Mixvpr: Feature mixing for visual
  place recognition. In: Proceedings of the IEEE/CVF Winter Conference on
  Applications of Computer Vision. pp. 2998--3007 (2023)

\bibitem{arandjelovic2016netvlad}
Arandjelovic, R., Gronat, P., Torii, A., Pajdla, T., Sivic, J.: Netvlad: Cnn
  architecture for weakly supervised place recognition. In: Proceedings of the
  IEEE Conference on Computer Vision and Pattern Recognition. pp. 5297--5307
  (2016)

\bibitem{arandjelovic2013all}
Arandjelovic, R., Zisserman, A.: All about vlad. In: Proceedings of the IEEE
  conference on Computer Vision and Pattern Recognition. pp. 1578--1585 (2013)

\bibitem{berton2022rethinking}
Berton, G., Masone, C., Caputo, B.: Rethinking visual geo-localization for
  large-scale applications. In: Proceedings of the IEEE/CVF Conference on
  Computer Vision and Pattern Recognition. pp. 4878--4888 (2022)

\bibitem{berton2023eigenplaces}
Berton, G., Trivigno, G., Caputo, B., Masone, C.: Eigenplaces: Training
  viewpoint robust models for visual place recognition. In: Proceedings of the
  IEEE/CVF International Conference on Computer Vision. pp. 11080--11090 (2023)

\bibitem{Boutros2021ElasticFaceEM}
Boutros, F., Damer, N., Kirchbuchner, F., Kuijper, A.: Elasticface: Elastic
  margin loss for deep face recognition. 2022 IEEE/CVF Conference on Computer
  Vision and Pattern Recognition Workshops (CVPRW) pp. 1577--1586 (2021)

\bibitem{cadena2016past}
Cadena, C., Carlone, L., Carrillo, H., Latif, Y., Scaramuzza, D., Neira, J.,
  Reid, I., Leonard, J.J.: Past, present, and future of simultaneous
  localization and mapping: Toward the robust-perception age. IEEE Transactions
  on robotics  \textbf{32}(6),  1309--1332 (2016)

\bibitem{cakir2019deep}
Cakir, F., He, K., Xia, X., Kulis, B., Sclaroff, S.: Deep metric learning to
  rank. In: Proceedings of the IEEE/CVF Conference on Computer Vision and
  Pattern Recognition. pp. 1861--1870 (2019)

\bibitem{campos2021orb}
Campos, C., Elvira, R., Rodr{\'\i}guez, J.J.G., Montiel, J.M., Tard{\'o}s,
  J.D.: Orb-slam3: An accurate open-source library for visual,
  visual--inertial, and multimap slam. IEEE Transactions on Robotics
  \textbf{37}(6),  1874--1890 (2021)

\bibitem{cao2020unifying}
Cao, B., Araujo, A., Sim, J.: Unifying deep local and global features for image
  search. In: Computer Vision--ECCV 2020: 16th European Conference, Glasgow,
  UK, August 23--28, 2020, Proceedings, Part XX 16. pp. 726--743. Springer
  (2020)

\bibitem{chen2022deep}
Chen, W., Liu, Y., Wang, W., Bakker, E.M., Georgiou, T., Fieguth, P., Liu, L.,
  Lew, M.S.: Deep learning for instance retrieval: A survey. IEEE Transactions
  on Pattern Analysis and Machine Intelligence  (2022)

\bibitem{cummins2008fab}
Cummins, M., Newman, P.: Fab-map: Probabilistic localization and mapping in the
  space of appearance. The International Journal of Robotics Research
  \textbf{27}(6),  647--665 (2008)

\bibitem{Deng2018ArcFaceAA}
Deng, J., Guo, J., Zafeiriou, S.: Arcface: Additive angular margin loss for
  deep face recognition. 2019 IEEE/CVF Conference on Computer Vision and
  Pattern Recognition (CVPR) pp. 4685--4694 (2018)

\bibitem{doan2019scalable}
Doan, A.D., Latif, Y., Chin, T.J., Liu, Y., Do, T.T., Reid, I.: Scalable place
  recognition under appearance change for autonomous driving. In: Proceedings
  of the IEEE/CVF International Conference on Computer Vision. pp. 9319--9328
  (2019)

\bibitem{galvez2012bags}
G{\'a}lvez-L{\'o}pez, D., Tardos, J.D.: Bags of binary words for fast place
  recognition in image sequences. IEEE Transactions on Robotics
  \textbf{28}(5),  1188--1197 (2012)

\bibitem{garcia2017hierarchical}
Garcia-Fidalgo, E., Ortiz, A.: Hierarchical place recognition for topological
  mapping. IEEE Transactions on Robotics  \textbf{33}(5),  1061--1074 (2017)

\bibitem{garg2021your}
Garg, S., Fischer, T., Milford, M.: Where is your place, visual place
  recognition? arXiv preprint arXiv:2103.06443  (2021)

\bibitem{ge2020self}
Ge, Y., Wang, H., Zhu, F., Zhao, R., Li, H.: Self-supervising fine-grained
  region similarities for large-scale image localization. In: Computer
  Vision--ECCV 2020: 16th European Conference, Glasgow, UK, August 23--28,
  2020, Proceedings, Part IV 16. pp. 369--386. Springer (2020)

\bibitem{hadsell2006dimensionality}
Hadsell, R., Chopra, S., LeCun, Y.: Dimensionality reduction by learning an
  invariant mapping. In: 2006 IEEE Computer Society Conference on Computer
  Vision and Pattern Recognition (CVPR'06). vol.~2, pp. 1735--1742. IEEE (2006)

\bibitem{harwood2017smart}
Harwood, B., Kumar~BG, V., Carneiro, G., Reid, I., Drummond, T.: Smart mining
  for deep metric learning. In: Proceedings of the IEEE International
  Conference on Computer Vision. pp. 2821--2829 (2017)

\bibitem{hausler2021patch}
Hausler, S., Garg, S., Xu, M., Milford, M., Fischer, T.: Patch-netvlad:
  Multi-scale fusion of locally-global descriptors for place recognition. In:
  Proceedings of the IEEE/CVF Conference on Computer Vision and Pattern
  Recognition. pp. 14141--14152 (2021)

\bibitem{hermans2017defense}
Hermans, A., Beyer, L., Leibe, B.: In defense of the triplet loss for person
  re-identification. arXiv preprint arXiv:1703.07737  (2017)

\bibitem{izquierdo2024optimal}
Izquierdo, S., Civera, J.: Optimal transport aggregation for visual place
  recognition. In: IEEE/CVF Conference on Computer Vision and Pattern
  Recognition (2024)

\bibitem{jegou2010aggregating}
J{\'e}gou, H., Douze, M., Schmid, C., P{\'e}rez, P.: Aggregating local
  descriptors into a compact image representation. In: 2010 IEEE Computer
  Society Conference on Computer Vision and Pattern Recognition. pp.
  3304--3311. IEEE (2010)

\bibitem{jin2017learned}
Jin~Kim, H., Dunn, E., Frahm, J.M.: Learned contextual feature reweighting for
  image geo-localization. In: Proceedings of the IEEE Conference on Computer
  Vision and Pattern Recognition. pp. 2136--2145 (2017)

\bibitem{kalantidis2020hard}
Kalantidis, Y., Sariyildiz, M.B., Pion, N., Weinzaepfel, P., Larlus, D.: Hard
  negative mixing for contrastive learning. Advances in Neural Information
  Processing Systems  \textbf{33},  21798--21809 (2020)

\bibitem{keetha2023anyloc}
Keetha, N., Mishra, A., Karhade, J., Jatavallabhula, K.M., Scherer, S.,
  Krishna, M., Garg, S.: Anyloc: Towards universal visual place recognition.
  IEEE Robotics and Automation Letters  (2023)

\bibitem{leyva2023data}
Leyva-Vallina, M., Strisciuglio, N., Petkov, N.: Data-efficient large scale
  place recognition with graded similarity supervision. In: Proceedings of the
  IEEE/CVF Conference on Computer Vision and Pattern Recognition. pp.
  23487--23496 (2023)

\bibitem{lowry2015visual}
Lowry, S., S{\"u}nderhauf, N., Newman, P., Leonard, J.J., Cox, D., Corke, P.,
  Milford, M.J.: {Visual Place Recognition: A Survey}. IEEE Transactions on
  Robotics  \textbf{32}(1),  1--19 (2015)

\bibitem{lu2023aanet}
Lu, F., Zhang, L., Dong, S., Chen, B., Yuan, C.: Aanet: Aggregation and
  alignment network with semi-hard positive sample mining for hierarchical
  place recognition. In: 2023 IEEE International Conference on Robotics and
  Automation (ICRA). pp. 11771--11778. IEEE (2023)

\bibitem{lu2024towards}
Lu, F., Zhang, L., Lan, X., Dong, S., Wang, Y., Yuan, C.: Towards seamless
  adaptation of pre-trained models for visual place recognition. International
  Conference on Learning Representations  (2024)

\bibitem{masone2021survey}
Masone, C., Caputo, B.: A survey on deep visual place recognition. IEEE Access
  \textbf{9},  19516--19547 (2021)

\bibitem{musgrave2020metric}
Musgrave, K., Belongie, S., Lim, S.N.: A metric learning reality check. In:
  Computer Vision--ECCV 2020: 16th European Conference, Glasgow, UK, August
  23--28, 2020, Proceedings, Part XXV 16. pp. 681--699. Springer (2020)

\bibitem{nordland}
NRK: Nordlandsbanen: minute by minute, season by season (2013),
  \url{https://nrkbeta.no/2013/01/15/nordlandsbanen-minute-by-minute-season-by-season/}

\bibitem{panek2023visual}
Panek, V., Kukelova, Z., Sattler, T.: Visual localization using imperfect 3d
  models from the internet. In: Proceedings of the IEEE/CVF Conference on
  Computer Vision and Pattern Recognition. pp. 13175--13186 (2023)

\bibitem{perronnin2010large}
Perronnin, F., Liu, Y., S{\'a}nchez, J., Poirier, H.: Large-scale image
  retrieval with compressed fisher vectors. In: 2010 IEEE Computer Society
  Conference on Computer Vision and Pattern Recognition. pp. 3384--3391. IEEE
  (2010)

\bibitem{radenovic2016cnn}
Radenovi{\'c}, F., Tolias, G., Chum, O.: Cnn image retrieval learns from bow:
  Unsupervised fine-tuning with hard examples. In: Computer Vision--ECCV 2016:
  14th European Conference, Amsterdam, The Netherlands, October 11--14, 2016,
  Proceedings, Part I 14. pp. 3--20. Springer (2016)

\bibitem{radenovic2018fine}
Radenovi{\'c}, F., Tolias, G., Chum, O.: Fine-tuning cnn image retrieval with
  no human annotation. IEEE Transactions on Pattern Analysis and Machine
  Intelligence  \textbf{41}(7),  1655--1668 (2018)

\bibitem{roth2020revisiting}
Roth, K., Milbich, T., Sinha, S., Gupta, P., Ommer, B., Cohen, J.P.: Revisiting
  training strategies and generalization performance in deep metric learning.
  In: International Conference on Machine Learning. pp. 8242--8252. PMLR (2020)

\bibitem{sarlin2021back}
Sarlin, P.E., Unagar, A., Larsson, M., Germain, H., Toft, C., Larsson, V.,
  Pollefeys, M., Lepetit, V., Hammarstrand, L., Kahl, F., et~al.: Back to the
  feature: Learning robust camera localization from pixels to pose. In:
  Proceedings of the IEEE/CVF Conference on Computer Vision and Pattern
  Recognition. pp. 3247--3257 (2021)

\bibitem{schroff2015facenet}
Schroff, F., Kalenichenko, D., Philbin, J.: Facenet: A unified embedding for
  face recognition and clustering. In: Proceedings of the IEEE Conference on
  Computer Vision and Pattern Recognition. pp. 815--823 (2015)

\bibitem{Schubert2023vpr}
Schubert, S., Neubert, P., Garg, S., Milford, M., Fischer, T.: {Visual Place
  Recognition: A Tutorial}. IEEE Robotics \& Automation Magazine  (2023)

\bibitem{shao2023global}
Shao, S., Chen, K., Karpur, A., Cui, Q., Araujo, A., Cao, B.: Global features
  are all you need for image retrieval and reranking. In: Proceedings of the
  IEEE/CVF International Conference on Computer Vision. pp. 11036--11046 (2023)

\bibitem{shen2023structvpr}
Shen, Y., Zhou, S., Fu, J., Wang, R., Chen, S., Zheng, N.: Structvpr: Distill
  structural knowledge with weighting samples for visual place recognition. In:
  Proceedings of the IEEE/CVF Conference on Computer Vision and Pattern
  Recognition. pp. 11217--11226 (2023)

\bibitem{simo2014fracking}
Simo-Serra, E., Trulls, E., Ferraz, L., Kokkinos, I., Moreno-Noguer, F.:
  Fracking deep convolutional image descriptors. arXiv preprint arXiv:1412.6537
   (2014)

\bibitem{smirnov2017doppelganger}
Smirnov, E., Melnikov, A., Novoselov, S., Luckyanets, E., Lavrentyeva, G.:
  Doppelganger mining for face representation learning. In: Proceedings of the
  IEEE International Conference on Computer Vision Workshops. pp. 1916--1923
  (2017)

\bibitem{sohn2016advances}
Sohn, K.: Improved deep metric learning with multi-class n-pair loss objective.
  Advances in Neural Information Processing Systems  (2016)

\bibitem{suh2019stochastic}
Suh, Y., Han, B., Kim, W., Lee, K.M.: Stochastic class-based hard example
  mining for deep metric learning. In: Proceedings of the IEEE/CVF Conference
  on Computer Vision and Pattern Recognition. pp. 7251--7259 (2019)

\bibitem{Sun2020CircleLA}
Sun, Y., Cheng, C., Zhang, Y., Zhang, C., Zheng, L., Wang, Z., Wei, Y.: Circle
  loss: A unified perspective of pair similarity optimization. 2020 IEEE/CVF
  Conference on Computer Vision and Pattern Recognition (CVPR) pp. 6397--6406
  (2020)

\bibitem{sunderhauf2015performance}
S{\"u}nderhauf, N., Shirazi, S., Dayoub, F., Upcroft, B., Milford, M.: On the
  performance of convnet features for place recognition. In: 2015 IEEE/RSJ
  International Conference on Intelligent Robots and Systems (IROS). pp.
  4297--4304. IEEE (2015)

\bibitem{sunderhauf2015place}
S{\"u}nderhauf, N., Shirazi, S., Jacobson, A., Dayoub, F., Pepperell, E.,
  Upcroft, B., Milford, M.: Place recognition with convnet landmarks:
  Viewpoint-robust, condition-robust, training-free. Robotics: Science and
  Systems XI pp. 1--10 (2015)

\bibitem{tomita2006worst}
Tomita, E., Tanaka, A., Takahashi, H.: The worst-case time complexity for
  generating all maximal cliques and computational experiments. Theoretical
  computer science  \textbf{363}(1),  28--42 (2006)

\bibitem{torii2013visual}
Torii, A., Sivic, J., Pajdla, T., Okutomi, M.: Visual place recognition with
  repetitive structures. In: Proceedings of the IEEE Conference on Computer
  Vision and Pattern Recognition. pp. 883--890 (2013)

\bibitem{wang2017deep}
Wang, J., Zhou, F., Wen, S., Liu, X., Lin, Y.: Deep metric learning with
  angular loss. In: Proceedings of the IEEE International Conference on
  Computer Vision. pp. 2593--2601 (2017)

\bibitem{wang2022transvpr}
Wang, R., Shen, Y., Zuo, W., Zhou, S., Zheng, N.: Transvpr: Transformer-based
  place recognition with multi-level attention aggregation. In: Proceedings of
  the IEEE/CVF Conference on Computer Vision and Pattern Recognition. pp.
  13648--13657 (2022)

\bibitem{wang2019multi}
Wang, X., Han, X., Huang, W., Dong, D., Scott, M.R.: Multi-similarity loss with
  general pair weighting for deep metric learning. In: Proceedings of the
  IEEE/CVF Conference on Computer Vision and Pattern Recognition. pp.
  5022--5030 (2019)

\bibitem{warburg2020mapillary}
Warburg, F., Hauberg, S., Lopez-Antequera, M., Gargallo, P., Kuang, Y., Civera,
  J.: Mapillary street-level sequences: A dataset for lifelong place
  recognition. In: Proceedings of the IEEE/CVF Conference on Computer Vision
  and Pattern Recognition. pp. 2626--2635 (2020)

\bibitem{weinberger2005distance}
Weinberger, K.Q., Blitzer, J., Saul, L.: Distance metric learning for large
  margin nearest neighbor classification. Advances in Neural Information
  Processing Systems  \textbf{18} (2005)

\bibitem{wu2017sampling}
Wu, C.Y., Manmatha, R., Smola, A.J., Krahenbuhl, P.: Sampling matters in deep
  embedding learning. In: Proceedings of the IEEE International Conference on
  Computer Vision. pp. 2840--2848 (2017)

\bibitem{xuan2020hard}
Xuan, H., Stylianou, A., Liu, X., Pless, R.: Hard negative examples are hard,
  but useful. In: Computer Vision--ECCV 2020: 16th European Conference,
  Glasgow, UK, August 23--28, 2020, Proceedings, Part XIV 16. pp. 126--142.
  Springer (2020)

\bibitem{yuan2019signal}
Yuan, T., Deng, W., Tang, J., Tang, Y., Chen, B.: Signal-to-noise ratio: A
  robust distance metric for deep metric learning. In: Proceedings of the
  IEEE/CVF Conference on Computer Vision and Pattern Recognition. pp.
  4815--4824 (2019)

\bibitem{yuan2017hard}
Yuan, Y., Yang, K., Zhang, C.: {Hard-Aware Deeply Cascaded Embedding}. In:
  Proceedings of the IEEE International Conference on Computer Vision. pp.
  814--823 (2017)

\bibitem{zhang2021visual}
Zhang, X., Wang, L., Su, Y.: {Visual Place Recognition: A Survey from Deep
  Learning Perspective}. Pattern Recognition  \textbf{113},  107760 (2021)

\bibitem{zhu2023r2former}
Zhu, S., Yang, L., Chen, C., Shah, M., Shen, X., Wang, H.: {R2Former: Unified
  Retrieval and Reranking Transformer for Place Recognition}. In: Proceedings
  of the IEEE/CVF Conference on Computer Vision and Pattern Recognition. pp.
  19370--19380 (2023)

\end{thebibliography}
